\documentclass{article}

\usepackage{arxiv}

\usepackage[utf8]{inputenc} % allow utf-8 input
\usepackage[T1]{fontenc}    % use 8-bit T1 fonts
\usepackage{hyperref}       % hyperlinks
\usepackage{url}            % simple URL typesetting
\usepackage{booktabs}       % professional-quality tables
\usepackage{amsfonts}       % blackboard math symbols
\usepackage{nicefrac}       % compact symbols for 1/2, etc.
\usepackage{microtype}      % microtypography
\usepackage{lipsum}
\usepackage{graphicx}
\usepackage{multicol}
\usepackage{booktabs}
\usepackage{multirow}
\usepackage{amssymb}
\usepackage{rotating}
\usepackage[dvipsnames]{xcolor}
\colorlet{LightLavender}{Lavender!40!}
\colorlet{LightSpringGreen}{SpringGreen!80!}
\usepackage{tcolorbox}
\usepackage[justification=centering]{caption}
\graphicspath{ {./images/} }

\title{Survey for Categorising Explainable AI Studies Using Data Analysis Task Frameworks}

\author{
 Hamzah Ziadeh \\
  Department of Design, Architecture, and Media technology\\
  Aalborg University\\
  Aalborg, Denmark \\
  \texttt{hazi@create.aau.dk} \\
   \And
 Hendrik Knoche \\
  Department of Design, Architecture, and Media technology\\
  Aalborg University\\
  Aalborg, Denmark \\
  \texttt{hk@create.aau.dk} \\
}
\begin{document}
\maketitle
\begin{abstract}
Research into explainable artificial intelligence (XAI) for data analysis tasks suffer from a large number of contradictions and lack of concrete design recommendations stemming from gaps in understanding the tasks that require AI assistance. In this paper, we drew on multiple fields such as visual analytics, cognition, and dashboard design to propose a method for categorising and comparing XAI studies under three dimensions: what, why, and who. We identified the main problems as: inadequate descriptions of tasks, context-free studies, and insufficient testing with target users. We propose that studies should specifically report on their users' domain, AI, and data analysis expertise to illustrate the generalisability of their findings. We also propose study guidelines for designing and reporting XAI tasks to improve the XAI community's ability to parse the rapidly growing field. We hope that our contribution can help researchers and designers better identify which studies are most relevant to their work, what gaps exist in the research, and how to handle contradictory results regarding XAI design.
\end{abstract}

% keywords can be removed
%\keywords{First keyword \and Second keyword \and More}

\section{Introduction}
Artificial intelligence (AI) can assist users with completing mentally demanding tasks such as data analysis. When analysing data, users need to make multiple complex decisions, such as what to analyse, methods to use, and how to interpret results~\cite{Sacha2016}. Furthermore, many analytical goals (e.g. predicting a patient's outcome) require users to keep track of results from prior analyses, understand uncertainty in data, and account for external factors (e.g. guidelines)~\cite{Brown2019, Sacha2016, Liu2020}. Therefore, AI systems aim to simplify parts of analysis tasks to curb mental demands. AI systems can assist in many ways, from recommending analytical actions and conclusions to automating analysis completely. While AI systems can completely replace users, most tasks need human input and oversight as AI systems' perspectives only consider the data they have access to. Many users (e.g. clinicians, teachers, etc) need control over AI systems and final decision making as they have responsibility over stakeholders~\cite{Correll2019}. Therefore, AI systems should only assist users with difficult steps in a task while maintaining users' sense of agency; their feeling of control over a task and its outcomes~\cite{Correll2019, Sun2019}.

Explainable AI (XAI) research investigates how models should explain themselves to users. This includes both explaining the inner workings of AIs models to providing evidence that supports a recommendation~\cite{Mohseni2021}. Current research has identified a large range of XAI methods (e.g. data visualisations, counterfactuals, etc) that can provide users with crucial information when deciding whether or not to trust a recommendation~\cite{Mohseni2021, Wang2021}. However, adding more explainability to an AI system does not always increase explainability as users may not understand explanations or become too overwhelmed with information~\cite{Wang2021, Sivaraman2023}. Therefore, XAI systems need to provide users with the most important explanations for their context and simplify those explanations to match users' familiarity with the data and AI model.  

XAI systems designers therefore need to understand tasks' motivations, goals, steps, and difficulties as well as users' knowledge and struggles. For this reason, many studies investigated AI support in analysis tasks in both lab~\cite{Sivaraman2023, Salimzadeh2024, Fast2018} and real world settings~\cite{Bach2023, Wang2021}. However, many studies found contradicting results regarding AI systems' benefits, shortcomings, and their effect on users. For example, many studies provided evidence for automation bias; users opting to always trust AI recommendations over their own judgements~\cite{Mosier1999, Goddard2014, Vered2023}. This lead to incorrect judgements in cases where AI systems provided incorrect recommendations. However, other studies found little to no evidence of automation bias~\cite{Salimzadeh2024, DeArteaga2020}. Even studies finding evidence often found different levels and  causes for it. The still limited studies on XAI systems and the increasing complexity of AI models further increase the contradictions in research. Many studies attempting to define guidelines for AI assistance struggled to create concrete recommendations for XAI design~\cite{Correll2019, Dudley2018, He2023, Tamblyn2008}. While many guideline papers outline design principles (e.g. consider user goals)~\cite{Mackeprang2019, Mohseni2021, Dudley2018}, guidelines often remain too high-level and provide little actionable advice (e.g. what specific XAI features should change based on user goals and how).
%Researchers
%Regulators
%Designers

In this paper, we hypothesise that the large number of contradictions in research into XAI for data analysis tasks and lack of concrete design recommendations stem from gaps in understanding analysis tasks that require AI assistance. To address these gaps and contradictions in research, we draw on literature from the fields of XAI, data analysis processes, data analysis tasks in real life (e.g. for healthcare), and human decision-making. In this systematic review, we propose a categorisation of data analysis tasks and use our method to unravel existing contradictions in XAI research. We hope our contribution will  support both researchers and XAI designers and researchers in better defining the tasks they investigate and designers in  data analysis to more easily recognise, which findings and guidelines apply in a given case.

%Using this classification method, we also propose more concrete guidelines to design AI assistance for data analysis. 

\section{Background}
When assisting users, XAI systems impact users in many ways, including their mental workload, trust, and sense of agency~\cite{Sivaraman2023, Wang2021, Calisto2022, Fast2018}. Domain experts (i.e. with expertise within a certain field such as lawyers and clinicians) need to experience a strong sense of agency as previously mentioned~\cite{sacha2015role}. However, studies have not distinguished between a preference for agency during the process of analysis and agency in the final decision-making (e.g. what treatment to give a patient). This can lead to difficulty in understanding how much control and customisation systems should provide (e.g. should users have the ability to include or exclude predictors from a model). 
While a strong sense of agency can enhance users' trust in a system and its recommendations, it may lead to customisation bias —- a tendency to over-trust the AI because users mistakenly believe that outputs reflect their own knowledge or input, rather than the model’s processing.

~\cite{Solomon2014}. Similarly, designs prioritizing the sense of agency can increase mental workload (i.e. the mental resources required)~\cite{Bach2023, Bird2012, Crisan2021}. The more control and insight a system provides, the more work users have to carry out in data analysis, which can overwhelm them~\cite{Sivaraman2023, Panigutti2022}.  Overloaded users either exhibit automation bias~\cite{Solomon2014, Wang2021, Salimzadeh2024} to avoid the mental demands of the task or avoid using the system due to its complicating of the task rather than simplifying it; especially when the AI and its explainability features increase in complexity~\cite{Sivaraman2023, Salimzadeh2024}. However, we currently do not understand why some users exhibit automation bias while others avoid using AI. 

Designers of AI systems need to balance these trade-offs to optimize workflows. However, many studies contradict each others' results on what kind of explainability users need. For example,  explainability increased users' confidence in their explanations~\cite{Lai2022, Panigutti2022} while the same explanations in a different system elevated doubt in their conclusions~\cite{Sivaraman2023}. While studies offered potential explanations these contradictions (e.g. lab setting v.s. in-situ)~\cite{Wang2024}, they lacked empirical evidence. In this section, we describe the overall context of data analysis and  the main relevant concepts related to XAI systems explored in previous studies.

\subsection{Data Use in Real Life}
The scope of this paper only considers externally motivated data analysis in which users make decisions and action plans about other people or their overall environment (e.g. teachers identifying, which students need extra attention)~\cite{Sun2019, Brown2019}. These users 
% performing these analyses 
have access to large data sets consisting of multiple variables for every item in the dataset (e.g. a patient has measurements of blood pressure, glucose, age, prescriptions, etc.)~\cite{Brown2019, Sparring2018}. Users need sufficient domain knowledge to understand the meaning of variables and how to use them~\cite{sacha2015role, WeggelaarJansen2018}. The size and complexity of datasets increase the difficulty of analysis as they require sufficient domain, analytical and data exploration knowledge~\cite{sacha2015role}. Typically, this data is available in tabular format and sometimes represented in dashboards that contain simple visualisation that give users initial insights for their analysis~\cite{Sparring2018, WeggelaarJansen2018, Sun2019}. The data can be multimodal and include pieces of text, timelines, images, or videos depending on the domain~\cite{DeArteaga2020, Bach2023, DeCroon2021}.

The granularity of data analysis can differ depending on users' goals and roles. For example, clinicians can evaluate data from a large number of patients to evaluate overall care quality or analyse data from a single patient to decide on a diagnosis~\cite{Rabiei2022}. Typically, users analysing data from aggregates or registries face more difficulties as the larger data sets come with increased complexity~\cite{Brown2019, WeggelaarJansen2018, Alhamadi2022}.

%The granularity of data analysis increases the analysis complexity and goals. 

\subsection{Decision and Sense Making}
\begin{figure}
    \centering
    \includegraphics[width=0.9\linewidth]{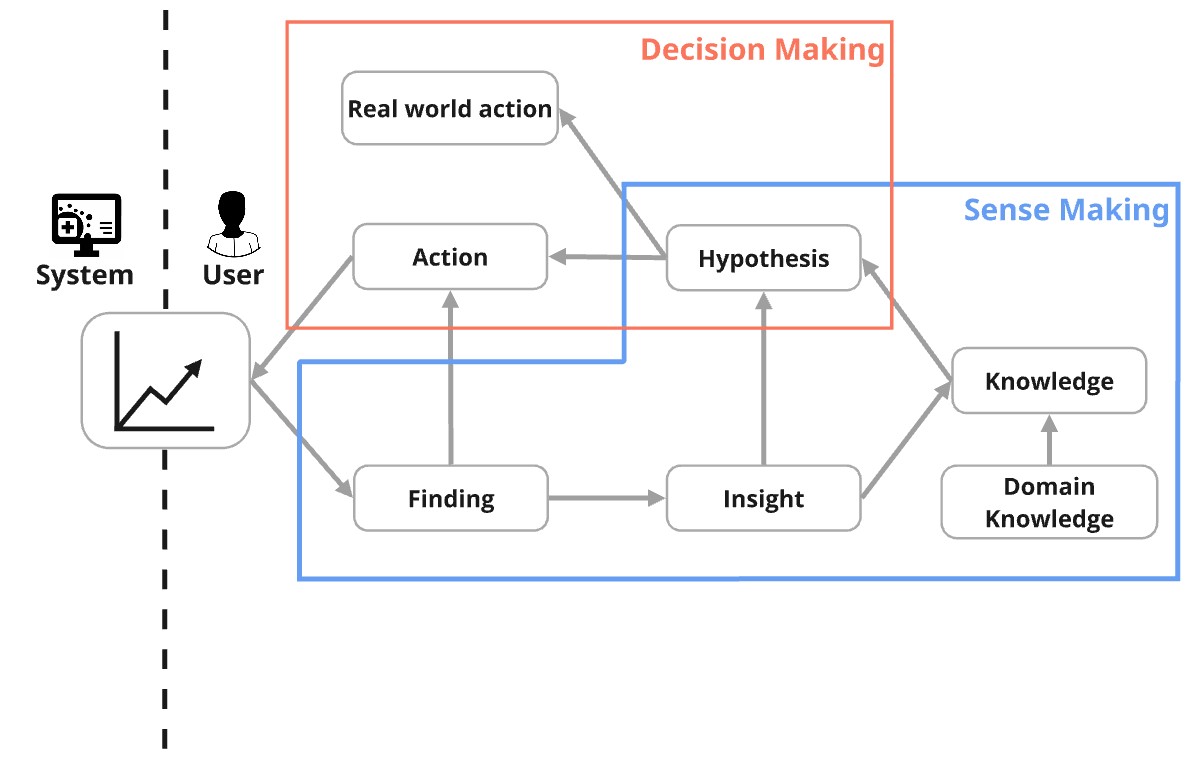}
    \caption{Knowledge Generation loop modified with domain knowledge, real world actions, and decision/sense making.}
    \label{fig:KGL}
\end{figure}

When analysing data, users switch back and forth between decision and sense making. The latter describes understanding the data and formulating new knowledge based on the insights gained~\cite{Dimara2022, WhitelockWainwright2022}. It relies on users' goals and domain knowledge, which in turn informs their decision making. Decision making includes both analytical decisions (e.g. what visualisation to make) and final decisions to act on in the real world (e.g. what medication to give a patient)~\cite{Dimara2022, sacha2015role}. To further describe the data analysis process, we integrated sacha et al.'s knowledge generation loop~\cite{sacha2015role} in which expert users begin data exploration with a \textit{hypothesis} about a problem (e.g. low care quality) using their domain knowledge. They then perform \textit{actions} to search for \textit{findings} (e.g. visualisations aggregating data) in a system and interpret them to extract \textit{insights} about their hypothesis (e.g. indicators reducing care quality). These insights add to their knowledge to generate new hypotheses. See Figure~\ref{fig:KGL} for the full process. While most users begin with a hypothesis, some users can begin analysis by looking at a dashboard or pre-made visualisations and searching for a finding that prompts further analysis (e.g. a patient unexpectedly getting worse).

\subsection{Automation}
Parasuraman et al. identified 10 degrees of AI task automation (see Table~\ref{Tab: loa}, from giving users complete control to AI having complete control (including deciding on the real world action)~\cite{Stumpf2009, Mackeprang2019}. Besides the automation level, AI can range from automating the entire task to only automating a single step~\cite{Mackeprang2019}. In some cases, AI systems can assist users in multiple steps with different automation levels. For example, a system can autonomously complete an action the users decided (level 10) and then only suggest a possible insight that users can get from the data (level 4).

Many AI systems automated the entire process of analysing data (e.g. finding students in need of assistance), which frees users to concentrate on sense and decision making (why they need help and what help to give)~\cite{Srinivasan2021, Salimzadeh2024, Wang2021}. These systems required users only to decide if they agreed with a recommendation or not (level 4)~\cite{Salimzadeh2024, Wang2021, Calisto2022, DeArteaga2020}. However, none of these studies investigated what users wanted and posited that users wanted recommendations after the analysis had completed automatically. This goes against recommendations that designers and researchers should begin with identifying which parts of a task need automation and assign appropriate automation levels for each~\cite{Stumpf2009,Mackeprang2019}. Therefore, these studies only investigated systems automating tasks at levels~4 and~5. Despite using similar levels, studies still found contradicting results. For example, studies in automation level~4 still did not agree on whether domain experts exhibited automation bias~\cite{Bach2023} or not~\cite{DeArteaga2020}. 

\begin{table*}[ht]
\centering
\renewcommand{\arraystretch}{1.2}
\begin{tabular}{@{}p{0.47\textwidth}@{\hspace{0.02\textwidth}}|@{\hspace{0.02\textwidth}}p{0.47\textwidth}@{}}
\multicolumn{2}{c}{\textbf{Levels of Automation}} \\[1ex]
\textbf{1.} The computer offers no assistance; the human must take all decisions and actions. &
\textbf{6.} The computer allows the human a restricted time to veto before automatic execution. \\

\textbf{2.} The computer offers a complete set of decision/action alternatives. &
\textbf{7.} The computer executes automatically, then necessarily informs the human. \\

\textbf{3.} The computer narrows the selection down to a few. &
\textbf{8.} The computer informs the human only if asked. \\

\textbf{4.} The computer suggests one alternative. &
\textbf{9.} The computer informs the human only if it, the computer, decides to. \\

\textbf{5.} The computer executes that suggestion if the human approves. &
\textbf{10.} The computer decides everything and acts autonomously, ignoring the human. \\
\end{tabular}
\caption{Descriptions of the 10 levels of automation as described by Parasuraman et al.~\cite{Parasuraman2000}.}
\label{Tab: loa}
\end{table*}

\subsection{Explainable AI}
XAI refers to AI systems that explain their inner working or reasoning for recommendations~\cite{Mohseni2021, Shneiderman2020}. XAI systems can present information and evidence to support a prediction or recommendations (local explanations) as well explain how the AI works (global explanation). For example, local explanations can include feature contribution (what specific data points contributed most to a prediction and how much)~\cite{Wang2021}. Global explanations can instead point to feature importance (the coefficient each variable has in the model)~\cite{Wang2021}. 

Local explanations aim to help users decide whether or not to trust a recommendation and provide relevant insights as evidence~\cite{Srinivasan2021, Wang2021, Calisto2022}. Local explanations can include counterfactuals (showing evidence contrary to the recommendation) or just presenting the data that the recommendation was based on (e.g. the patient data that yielded the recommendation for a treatment)~\cite{Wang2021}. Almost all XAI studies included a form of local explanations as users need them to decide whether to accept or reject a recommendation~\cite{Sivaraman2023}. But they disagree on which local explanations to include in their interfaces. For example, some studies found that counterfactuals increased trust in a system as users felt better informed and spent more time considering the recommendation~\cite{Dai2022, Bach2023} while others  found no such benefits and suggested that counterfactuals only increased mental workload~\cite{Wang2021}.

Global explanations can include model cards that describe the training data of a model, known biases, updates, and intended use~\cite{Mitchell2019, Dodge2019} to reduce misuse, inform user of AI's strengths and weaknesses, and increase trust in the overall model. But again, studies disagreed on their use. For example, some studies found that model cards increased user trust and reduced automation bias~\cite{Dodge2019} while others found that adding feature importance and model cards had no effect on users~\cite{Wang2021, Lai2022}.

\section{Conversational AI}
Many recent studies on AI assistance in data analysis investigated using conversational AI (CAI), which users interacted with by typing what actions they wanted to do or hypotheses they wanted information on~\cite{Fast2018, Srinivasan2021}. Following suggestions on non-AI assisted analysis tools (e.g. Tableau), many users preferred defining their goals and then working to achieve them top-down; deciding on the visualisation type and then editing it step by step to achieve what they want~\cite{Mndez2017}. studies supported this hypothesis by showing that many users experienced less mental workload analysing data with CAIs as opposed to traditional tools (Tableau, Jupyter, etc)~\cite{Fast2018, Srinivasan2021, Crovari2021}.

CAIs not only offer easier interactions, but allow users to define their goals and questions so that the system can better align itself with their needs. For example, users explaining what data they want to analyse and why so CAIs can limit the scope of analysis and recommendations to what users need~\cite{Crovari2021}. CAIs can  suggest analytical methods or  assist by generating code for analysis~\cite{Fast2018, Li2023}. CAI studies typically had their systems recommend analytical actions rather than automating the analysis and recommending conclusions and real world actions~\cite{Fast2018}. Most CAI systems  gave users multiple options to choose from (level~3)~\cite{Fast2018, Srinivasan2021}. 

Despite many studies showing the benefits of CAI, very few compared CAI designs to understand how to design them. For example, no study investigated whether a CAI should explain the reasons behind action recommendations. Similar to XAI studies, CAI studies still suffer from many contradictions. However, no studies so far have investigated how AI and CA systems can support such tasks. For example, when the CAI systems completed actions (e.g. normalising data) for users without explaining how it did it, some users trusted the CAI~\cite{Crovari2021} while others required more transparency (e.g. what  methods were used)~\cite{Fast2018}.

\section{Survey Method}
We surveyed existing research into both XAI and data analysis systems to capture a wide breadth of data analysis tasks, user needs, and AI assistance. We followed a three-step process for both understanding the current scope of AI assistance in analysis tasks and collect current XAI and CAI research to categorise their tasks and resolve identified contradictions.

We started by searching for existing survey and review papers related to data analysis tasks using \textit{Google Scholar}. To ensure a broad understanding of how people engage with data, we looked for studies, surveys, and reviews on sense making, exploratory data analysis, data visualisation, data dashboards, information visualisation, and visual analytics. This search resulted in 37 relevant papers and one book. We included only the papers that focused on identifying data analysis tasks and what struggles users faced during these tasks, which results in 13 survey papers, nine studies, and one book. 

In the second step, we searched for survey and review papers on AI assistance. Given the rapid development of this field, we included both peer-reviewed studies and arXiv preprints. We focused on papers covering explainability, interpretability, transparency, human-AI collaboration, human-in-the-loop AI, interactive AI, and decision support systems. This resulted in 55 studies. We included only those focusing on identifying tasks, types of AI assistance, and user needs in AI assistance. We excluded papers that focused on technical development of AI (e.g. reducing algorithmic bias in models). This resulted in 11 papers.

Finally, we further searched for studies evaluating AI systems in sense and decision making tasks using the same key words used in the previous step. This resulted in 144 papers. Once again, we excluded papers with a technical focus on included only those with a focus on user evaluations in relevant sense/decision making tasks. This resulted in a final 37 studies that we used to find contradictions and resolve them through categorising their tasks.

\section{Categorisation of XAI studies}
This section describes the characteristics we identified that can distinguish tasks between studies. Table~\ref{tab:sota-overview} shows all the studies collected in this review and which characteristics they fulfil. We categorised tasks across three dimensions: Who, Why, and What, which will be described below.

\begin{table}
\centering
\caption{Overview of studies included in this survey and a breakdown of their task type based on our framework. If a study investigated multiple tasks, the different tasks have seperate rows.}
\label{tab:sota-overview}
\resizebox{\textwidth}{!}{%
\begin{tabular}{cccccccccccc}
\hline
\multirow{3}{*}{ID} &
  \multirow{3}{*}{CXAI} &
  \multirow{3}{*}{\begin{tabular}[c]{@{}c@{}}Target \\ group\end{tabular}} &
  \multirow{3}{*}{\begin{tabular}[c]{@{}c@{}}Contextual \\ task\end{tabular}} &
  \multirow{3}{*}{\begin{tabular}[c]{@{}c@{}}User\\ expertise\end{tabular}} &
  \multirow{3}{*}{\begin{tabular}[c]{@{}c@{}}Sense making\\ type\end{tabular}} &
  \multirow{3}{*}{\begin{tabular}[c]{@{}c@{}}Decision making\\ type\end{tabular}} &
  \multirow{3}{*}{\begin{tabular}[c]{@{}c@{}}Prediction \\ type\end{tabular}} &
  \multirow{3}{*}{\begin{tabular}[c]{@{}c@{}}Decision \\ granularity\end{tabular}} &
  \multicolumn{3}{c}{Data type} \\ \cline{10-12} 
 &
   &
   &
   &
   &
   &
   &
   &
   &
  \multicolumn{2}{c}{Quant.} &
  \multirow{2}{*}{Qual.} \\ \cline{10-11}
 &
   &
   &
   &
   &
   &
   &
   &
   &
  Cat. &
  Ord. &
   \\ \hline
\cite{Bach2023} &
   &
  \checkmark &
  \checkmark &
  Domain &
  \multicolumn{1}{c}{D} &
  E &
  C &
  B &
   &
   &
  \checkmark \\
\cite{Calisto2022} &
   &
  \checkmark &
  \checkmark &
  Domain &
  D &
  E &
  C &
  B &
   &
   &
  \checkmark \\
\cite{Corti2024} &
   &
  \checkmark &
  \checkmark &
  Domain &
  D &
  E+C &
  C &
  B &
  \checkmark &
  \checkmark &
  \checkmark \\
\cite{Bach2023} &
   &
  \checkmark &
  \checkmark &
  Domain &
  D &
  C &
  C &
  B &
   &
   &
  \checkmark \\
\cite{Karagoz2024} &
   &
  \checkmark &
  \checkmark &
  Domain &
  D &
  C &
  C &
  B &
   &
   &
  \checkmark \\
\cite{Sivaraman2023} &
   &
  \checkmark &
  \checkmark &
  Domain &
  D &
  C &
  C &
  C &
  \checkmark &
  \checkmark &
   \\
\cite{Panigutti2022} &
   &
  \checkmark &
  \checkmark &
  Domain &
  D &
  C &
  C &
  B &
  \checkmark &
   &
   \\
\cite{Lundberg2018} &
   &
  \checkmark &
  \checkmark &
  Domain &
  P &
  E &
  O &
  B &
  \checkmark &
  \checkmark &
   \\
\cite{Zhang2024} &
   &
  \checkmark &
  \checkmark &
  Domain &
  P &
  E+C &
  O &
  B &
   &
  \checkmark &
   \\
\cite{Kaur2020} &
   &
   &
   &
  \begin{tabular}[c]{@{}c@{}}Analysis + \\ AI\end{tabular} &
  D &
  E &
  O &
  B &
  \checkmark &
  \checkmark &
   \\
\cite{Liebherr2025} &
   &
   &
   &
   &
  D &
  E &
  O &
  B &
  \checkmark &
  \checkmark &
  \checkmark \\
\cite{Lai2022} &
   &
   &
   &
   &
  D &
  E &
  C &
  B &
   &
   &
  \checkmark \\
\cite{Bansal2021} &
   &
   &
   &
   &
  D &
  E &
  C &
  B &
   &
   &
  \checkmark \\
\cite{Linder2021} &
   &
   &
   &
   &
  D &
  E &
  O &
  B+C &
   &
   &
  \checkmark \\
\cite{Salimzadeh2024} &
   &
   &
   &
   &
  D &
  E &
  O &
  C &
  \checkmark &
  \checkmark &
   \\
\cite{Wang2021} &
   &
   &
   &
   &
  D &
  E &
  C &
  B &
  \checkmark &
  \checkmark &
   \\
\cite{Buinca2021} &
   &
   &
   &
   &
  D &
  E &
  O &
  B &
   &
  \checkmark &
   \\
\cite{Salimzadeh2024} &
   &
   &
   &
   &
  P &
  E &
  C &
  B &
  \checkmark &
  \checkmark &
   \\
\cite{Dodge2019} &
   &
   &
   &
   &
  P &
  E &
  C &
  B &
  \checkmark &
  \checkmark &
   \\
\cite{Wang2021} &
   &
   &
   &
   &
  P &
  E &
  C &
  B &
  \checkmark &
  \checkmark &
   \\
\cite{Cau2023} &
   &
   &
   &
   &
  P &
  E &
  C &
  B &
   &
  \checkmark &
   \\
\cite{Ahn2024} &
   &
   &
   &
   &
  P &
  C &
  O &
  C &
  \checkmark &
  \checkmark &
   \\
\cite{Solomon2014} &
    &
   &
   &
   &
  P &
  C &
  C &
  B &
   &
  \checkmark &
   \\
\cite{Warren2024} &
   &
   &
   &
   &
  I &
  E &
  C &
  B &
  \checkmark &
  \checkmark &
   \\
\cite{Rajashekar2024} &
  \checkmark &
  \checkmark &
  \checkmark &
  Domain &
  P &
  E+C &
  O &
  B &
  \checkmark &
  \checkmark &
   \\
\cite{Kuang2024} &
  \checkmark &
  \checkmark &
  \checkmark &
  \begin{tabular}[c]{@{}c@{}}Domain + \\ Analysis\end{tabular} &
  I &
  E+C &
  C &
  C &
   &
   &
  \checkmark \\
\cite{Crovari2021} &
  \checkmark &
  \checkmark &
   &
  \begin{tabular}[c]{@{}c@{}}Domain + \\ Analysis\end{tabular} &
  D &
  E &
  C &
  B &
  \checkmark &
  \checkmark &
   \\
\cite{Fast2018} &
  \checkmark &
  \checkmark &
   &
  Analysis + AI &
  D &
  E &
  C &
  B &
  \checkmark &
  \checkmark &
   \\
\cite{Srinivasan2021} &
  \checkmark &
  \checkmark &
   &
  Analysis &
  D &
  E &
  C &
  B &
  \checkmark &
  \checkmark &
   \\
\cite{Zhao2022} &
  \checkmark &
  \checkmark &
   &
  Analysis &
  D &
  E &
  C &
  B &
  \checkmark &
  \checkmark &
   \\
\cite{Shin2023} &
  \checkmark &
  \checkmark &
   &
  Analysis &
  D &
  E &
  C &
  B &
  \checkmark &
  \checkmark &
   \\
\cite{Gu2024} &
  \checkmark &
  \checkmark &
   &
  Analysis &
  D &
  E &
  C &
  B &
  \checkmark &
  \checkmark &
   \\
\cite{Yu2020} &
  \checkmark &
  \checkmark &
   &
  Analysis + AI &
  I &
  E &
  C &
  B &
  \checkmark &
  \checkmark &
   \\
\cite{Jain2018} &
  \checkmark &
   &
   &
   &
  D &
  E &
  C &
  B &
  \checkmark &
  \checkmark &
   \\
\cite{Khurana2021} &
  \checkmark &
   &
   &
  Analysis + AI &
  I &
  E &
  C &
  B &
  \checkmark &
  \checkmark &
   \\
\cite{Li2023} &
  \checkmark &
   &
   &
  Analysis + AI &
  I &
  E &
  C &
  B &
  \checkmark &
  \checkmark &
   \\
\cite{Ruoff2023} &
  \checkmark &
   &
   &
   &
  I &
  E &
  C &
  B &
  \checkmark &
  \checkmark &
   \\
\cite{Lee2021} &
  \checkmark &
   &
   &
   &
  I &
  E &
  C &
  B &
  \checkmark &
  \checkmark &
   \\
\cite{He2023} &
  \checkmark &
   &
   &
   &
  I &
  E &
  C &
  B &
  \checkmark &
  \checkmark &
   \\
\cite{Ma2025} &
  \checkmark &
   &
   &
   &
  I &
  E &
  C &
  B &
  \checkmark &
  \checkmark &
   \\ \cline{1-10} \cline{10-12} 
\end{tabular}
}
\end{table}

\subsection{Who is analysing?}
Users can differ in their levels of knowledge across three categories: 1) Domain, 2) analysis, and 3) AI knowledge. Even domain experts can have varying degrees of knowledge depending on their experience. Higher domain knowledge leads to decreased reliance on local explanations as domain experts can make decisions just by looking at raw data (e.g. they can decide on a treatment just by looking at patient data)~\cite{Wang2021}. Therefore, domain experts rely more on global explanations as they value understanding how models work overall rather than understanding specific recommendations~\cite{Wang2021}. When using CAI, domain experts want brief and actionable responses without explaining domain concepts~\cite{Rajashekar2024, Wang2024}. Inversely, domain novices prefer CAI systems to explain domain concepts along with any insights from the data analysis~\cite{Wang2024}. Typically, domain experts also require having more control over CAI systems' analysis process as they want to fine tune analysis to search for specific insights~\cite{Li2023, Fast2018}.

Analysis knowledge refers to expertise in data analysis and statistical methods. Users with high analysis knowledge typically have pre-defined workflows they developed over their years of working and prefer to have control over analytical decision making (e.g. what statistical tests to use)~\cite{Liu2020, Fast2018, Srinivasan2021}. They mainly seek AI systems that can automate executing actions they have defined to time and effort~\cite{Fast2018, Srinivasan2021, Crovari2021}. Analysis experts can still benefit from recommendations of analytical actions to help them consider alternative methods~\cite{Srinivasan2021, Li2023, Shin2023}. Overall, analytical experts value the ability of defining the exact steps of an analysis and would not accept a system that automates the analysis and recommends a final decision~\cite{Fast2018, Shin2023, Li2023}.

%Finally, AI knowledge refers to users' knowledge of AI systems. 
Finally, we also distinguish AI knowledge as a seperate set of expertise. This refers to knowledge of AI models, how they work, and how explainability methods work~\cite{Kaur2020, Crisan2021}. We distinguish AI knowledge from domain knowledge as understanding how AI models work and their limitations can change how users rely on them compared to domain experts in eduction, medicine, or other fields~\cite{Crisan2021}. AI experts demand more specific information about the model training and testing performance (e.g. model cards) to understand how to use a model~\cite{Crisan2021, Fast2018}. Typically, AI experts do not demand the same explainability of CAI models, but when using CAI to build other AI models, AI experts want to control over every step in the process~\cite{Fast2018, Fast2018}. 

As seen in Table~\ref{tab:sota-overview}, only twelve of the 41 papers we reviewed conducted their studies with the domain experts that their systems targetted while other studies only recruited crowd workers or data analysts. This leaves a gap in understanding how domain experts without analytical expertise use AI systems and how we need to design AI assistance to target their needs.

\subsection{Why are they analysing?}

\subsubsection{Sense making type}
Previous XAI research split data analysis tasks into two categories: diagnostic or prognostic~\cite{Salimzadeh2024}. Diagnostic tasks have clear objectives, verifiable conclusions about the past or present (e.g. patient does or does not have cancer), and data sets that include almost all relevant variables. Prognostic tasks involve making predictions (e.g. will the patient recover within a year) using data that might not capture all relevant information (e.g. family support a patient has or adherence to treatment). Users typically seek the most AI help (both involving and excluding CAI) in prognostic tasks, which most studies consider as more difficult due to the increased uncertainty~\cite{Salimzadeh2024, Wang2021, Lundberg2018}. Some studies hypothesised that this increased AI reliance stems from the mental workload of factoring large amounts of data to predict a future outcome that may depend on factors not represented in data~\cite{Wang2021, Salimzadeh2024}.  

While the majority of AI assistance studies without CAI focus on prognostic tasks, CAI studies mainly investigated diagnostic tasks as seen in Table~\ref{tab:sota-overview}. However, we propose that many domains require analysis that fits into neither category as users need to explain past or current phenomena despite not having access to crucial variables. For example, clinicians analysing the care quality of their hospital may not have the necessary data to identify exactly why patient outcomes have worsened (e.g. overall public health, when patient realised they need assistance, etc)~\cite{WeggelaarJansen2018, Desveaux2021}. We refer to these analyses as inferential tasks. In the example of care quality, users can hypothesise why their hospital has shortcomings and support their conclusions with data but must still use their domain knowledge (e.g. medical knowledge and knowledge of the hospital) to infer further explanations~\cite{Desveaux2021, Rajashekar2024}. %This can explain why clinicians analysing their care quality often struggle to reach concrete conclusions and formulate action plans despite having access to very large datasets~\cite{Brown2019}. We hypothesise that CAs must also support these inferential tasks and have behaviour designed to specifically support such analysis. 

\subsubsection{Decision making type}
Previous research distinguished decision making in data analysis tasks between exploratory and confirmatory data analysis~\cite{Tukey1980, Schwab2020, Fife2022}. In exploratory analysis, users begin with little to no idea of what they might find in the data (i.e. they have no concrete hypothesis). In such analysis, users need to explore the data, discover findings, and formulate completely new hypothesis. The decision making mainly occurs in what insights to incorporate into their knowledge and formulating a hypothesis on the data~\cite{sacha2015role}. In confirmatory analysis, users begin with a hypothesis and search for findings that support or go against their current hypothesis~\cite{sacha2015role}. In such analysis, the decision making mainly occurs in updating their hypothesis~\cite{sacha2015role}. 

Most XAI and CXAI studies investigated exploratory data analysis as seen in Table~\ref{tab:sota-overview}, which leaves a gap in understanding confirmatory analysis. This goes against most use cases with domain experts who usually form opinions by looking at the raw data and need AI systems to confirm, expand, or alter their hypotheses~\cite{sacha2015role, Correll2019, Zhang2024}. Focusing on exploratory analysis may not reflect real use cases where domain experts have lots of information before even analysing the data (e.g. through spending time stakeholders, being in the environment, and having experience). Displaying recommendations before users formulate a hypothesis increases AI reliance~\cite{Bach2023} whereas allowing users to form hypotheses on the case and then confirming it with the help of AI can lead to more appropriate AI use as well as more accurate decision making~\cite{Bach2023, Sivaraman2023, Zhang2024}. Therefore, studies need to design their tasks to align with the real tasks that users have to perform. 
%For example, clinicians will often decide on a diagnosis based on the raw data and teachers will already know which of their students need assistance. This alters users' need for explainability. In exploratory analysis, users look at explainability to understand whether the presented 

\subsection{What are they analysing?}

\subsubsection{Quantitative or Qualitative}
The type of data users analyse also changes the type of AI assistance they need. In this paper, we split data types in two: data-driven and information-driven. In data-driven tasks, users analyse a structured dataset containing multiple variables (e.g. csv, rmd, registry data, etc)~\cite{Fast2018, Srinivasan2021, Zhang2024}. This can consist of numerical or categorical data that users can explore using visualisations and statistical models~\cite{Mndez2017}. AI assistance in data-driven tasks involve either recommending analytical decisions~\cite{Fast2018, Srinivasan2021} or real-world decisions~\cite{Zhao2022, Cau2023}. XAI in data driven-tasks should similarly include data, showing feature contribution for local explanations and feature importance for global explanation~\cite{Wang2024}.

Information-driven tasks include data in the form of text, images, or videos~\cite{Kuang2024, Bach2023, DeArteaga2020, Linder2021}. In such tasks, users need to search through this information to find something (e.g. an error in text or a tumour in a lung image). Since these tasks rely more on users' perceptual abilities, AI assistance comes in the form of highlighting or drawing attention towards points of interest (e.g. circling a tumour or highlighting an error a student made in an essay)~\cite{Karagoz2024, DeCroon2021, Linder2021}. Users can judge the accuracy of the recommendation just by looking at or reading the point of interest and making up their own mind~\cite{Karagoz2024, Calisto2022}. Therefore, many AI systems in information driven tasks do not include local explanations~\cite{Calisto2022, Qu2023, Karagoz2024, DeCroon2021}. However, the perceptual demands of information-driven tasks can lead to mental fatigue, prompting users to exhibit higher automation bias and not reviewing all the data in favour of only looking at information the AI has highlighted~\cite{Bach2023, Qu2023}. Therefore, information-driven AI assistance should also include global explanations so users know what information the AI can locate and what it will likely miss~\cite{Stumpf2009}. 

\subsubsection{Quantitative data types for predictors}
Specifically within quantitative data-driven tasks, we can distinguish between categorical data (e.g. gender) or Ordered data (e.g. age, price, length)~\cite{munzner2015visualization}. When reviewing the data behind an AI recommendation, categorical data leads to more accurate decision making as well as improved ability to detect wrong recommendations~\cite{Warren2024}. Therefore, some XAI studies transformed ordered data into categorical~\cite{Warren2024, Salimzadeh2024, Srinivasan2021} or presented explanations as if the data was categorical through rules (e.g. buy this stock because prices decreased below a set threshold)~\cite{Cau2023}.

\subsubsection{Data type of predicted value}
Similar to predictors, systems can also either predict a value for either a categorical or ordered variable. Most system predict categorical data~\cite{Bach2023, Wang2021, Panigutti2022}. For example, whether patient has a specific illness~\cite{Panigutti2022} or whether a student should be admitted to a study programme~\cite{Ma2025}. Predicting categorical values can either involve a binary decision between two categories where users only have to accept or reject a recommendation~\cite{Panigutti2022, Ma2025, Warren2024} or decide between multiple categories (e.g. which medication to give a patient~\cite{Srinivasan2021}.

Alternatively, systems can predict an ordered value such as stock price~\cite{Cau2023} or risk of complication during surgery~\cite{Zhang2024, Lundberg2018}. In such cases, users disagreeing with a prediction have to forecast their own value or a possible range for the value, which requires extensive domain knowledge~\cite{Zhang2024}. While no XAI studies to our knowledge compared categorical and ordered decisions, frameworks of mental workload explain that reducing the range of choices can lower the mental workload required to decide~\cite{Longo2022}. 

\subsubsection{Complexity}
Previous studies also classify tasks based on complexity; defined as the information load required by task~\cite{Wood1986}. XAI or data exploration studies operationalise complexity as the number of variables in a dataset (four = low complexity, eight = medium, twelve = high)~\cite{Rong2024, Parkes2016, Salimzadeh2024}. Previous studies found that increasing complexity leads to higher AI dependency during decision making as users seek to lower their mental workload~\cite{Parkes2016, Salimzadeh2024}. However, this complexity categorisation only reflects simpler tasks (e.g. planning a route for a trip or classifying an animal species) but does not reflect many datasets in high stakes domain (e.g. healthcare), which can have datasets with much more variables, sometimes even hundreds~\cite{Sparring2018, Brown2019, WeggelaarJansen2018}. However, we can classify such tasks as high complexity. Only on of the studies in Table~\ref{tab:sota-overview} reported on their task complexity~\cite{Salimzadeh2024} while all other studies did not report on the size of their datasets. This lack of reporting makes it difficult to compare the studies as we can not evaluate whether the task complexity could account for misaligned results. 

\subsubsection{Severity}
Finally, the severity of the case that users have to make a decision on can impact how users interact with explanations~\cite{Salimzadeh2024}. For example, when deciding on whether to admit a student to a study programme; students with exceptionally good or bad academic performance have clear choices~\cite{Ma2025}. In such cases with extreme low or high severity, users may even forego looking at explanation or even the prediction as the choice may seem obvious from just looking at the data~\cite{Salimzadeh2024}. However, users often rely more heavily on AI assistance and explanations when cases have more medium severity; that have large amounts of uncertainty, lack any extreme values to indicate a choice, or have conflicting data pointing towards opposite choices. In such medium severity cases, users want to explore explanations in hopes to find insights that identify a decision~\cite{Salimzadeh2024, Zhang2024, Bach2023}. While many studies found that severity does impact how often users rely on AI and process XAI explanations, most studies do not report on the distribution of the severity for the decisions that users had to make, which makes it difficult to use it as an explanation for why results between studies differ. Similar to complexity, only one study reported on case severity~\cite{Calisto2022}, furthering the difficulty of comparing studies.

\begin{table}[ht]
\centering
\caption{Three types of analysis tasks supported AI systems.}
\resizebox{\textwidth}{!}{%
    
    \begin{tabular}{p{1.5cm} p{2.5cm} p{2cm} p{2.8cm} p{2.5cm} p{5cm}}
    
    \textbf{Task Type} & \textbf{\begin{tabular}[c]{l} Data\\Completeness\end{tabular}} & \textbf{Time Focus} & \textbf{Uncertainty} & \textbf{ \begin{tabular}[c]{l} Domain\\knowledge\\Need\end{tabular}} & \textbf{Examples} \\
    \hline
    Prognostic & 
    Incomplete data & 
    Future & 
    High & 
    High & 
    \begin{tabular}[c]{l}Predicting patient response\\ to treatment. \\ Forecasting future hospital\\ performance.\end{tabular} 
    \\
    \\
    Diagnostic & 
    Most relevant data included & 
    Present & 
    Low & 
    Low & 
    \begin{tabular}[c]{l}Check if patient received\\ treatment following guidelines. \\ Compare hospital treatment \\ times to national aggregate.\end{tabular} 
    \\
    \\
    Inferential & 
    Incomplete data & 
    Present & 
    High & 
    Medium & 
    \begin{tabular}[c]{l}Explain unexpected change \\ in patient health. \\ Explain reason for lower \\ hospital performance\end{tabular} 
    \\
    \\
    \end{tabular}
 }
\label{tab:task-types}
\end{table}

\section{Reconciling Contradictory Results}
In this section, we cover three XAI topics that have misaligned results across studies: automation bias, effects of global explanations, and the level of explainability in CXAI systems. This section aims to showcase how to use our categorisation method to reconcile contradictory results in research and propose explanations for some of the contradictions we found. 

\subsection{Automation Bias}
From the studies we collected, nine found evidence of automation bias~\cite{Wang2021, Buinca2021, Solomon2014, Levy2021, Qu2023, Salimzadeh2024, Panigutti2022, Bach2023, Solomon2014} while four did not~\cite{Sivaraman2023, Calisto2022, Kuang2024, DeArteaga2020}. Five of the studies with automation bias conducted their studies on Amazon Turk participants and performed tasks that they did not have a stake or knowledge in (e.g. judging whether a criminal will reoffend)~\cite{Wang2021, Buinca2021, Bansal2021, Salimzadeh2024, Solomon2014}. This could impact results as Amazon turk participants aim to perform many tasks as fast as possible and thus do not put in as much effort compared to participants testing systems meant to help them at their jobs~\cite{Wang2024}. Two studies with automation bias did have domain experts (e.g. medical students) but tested systems that did not match any task they perform in their context (e.g. participants annotated medical articles with the topic they discussed)~\cite{Levy2021, Qu2023}. Previous studies on mental workload found that users performing tasks they have no motivation for (e.g. receiving no benefit from it or not relating to their job) often put less effort into the task and fall into satisficing; doing the bare minimum effort and not thoroughly thinking through choices~\cite{Longo2022}. The last three studies with automation did include both domain experts and a task that matches a real task they perform and found much weaker evidence for automation bias~\cite{Corti2024, Panigutti2022, Bach2023}. Similarly, the four studies without automation bias all had domain experts and tasks fitting their context~\cite{Sivaraman2023, Calisto2022, Kuang2024, DeArteaga2020}. This suggests that a large cause of automation bias in previous studies could stem from participants having no incentive to properly review AI recommendations. Most studies with automation bias also investigated prognostic tasks~\cite{Buinca2021, Wang2021, Salimzadeh2024, Solomon2014}, which also can increase automation bias due to the increased difficulty of the task~\cite{Salimzadeh2024}.

\subsection{Global Explanations}
% Studies disagreed whether XAI systems should improve on local explanations of their models by including global explanations. 
Studies disagreed whether XAI systems should include local explanations alone or be supplemented with global explanations. While global explanations helped users understand models more~\cite{Dodge2019, Wang2021}, increase trust~\cite{Wang2021}, and avoid over-trusting the model~\cite{Wang2021} in some studies they only increased workload without improving task performance~\cite{Ahn2024, Lai2022}. Studies arguing in favour of global explanations involved prognostic tasks where users needed explanations and decide if they agreed with a prediction (e.g. risk of criminal re-offence). Moreover, they included data-driven tasks where users valued understanding the effect of each variable on the outcome. On the other hand, studies showing no effect of global explanations had information driven prognostic tasks where users could more easily verify the explanations by looking at text and reaching a conclusion. 

This suggests that experts using XAI systems to make predictions want to understand the AI pipeline behind the predictions to decide on whether to trust systems. This specifically applies to data-driven tasks as the relationship between a prediction and the source data has little clarity compared to information-driven tasks (e.g. presence of keyword in a text determines the prediction). Similarly, many studies on information-driven tasks show that users do no ask for global explanations as they can review the source data (e.g. text, images, and videos) themselves to form an opinion and only use the XAI system as support to confirm their findings~\cite{Bach2023, Karagoz2024, Linder2021, Calisto2022}.

\subsection{Explainability in conversations}
Many studies on CXAI disagreed on whether users wanted systems to specify their analytical methods (e.g. which statistical tests were used)~\cite{Fast2018, Srinivasan2021, Rajashekar2024, Gu2024} or only provide results without specifying the methods~\cite{Ma2025,He2023}. Studies in support of adding such specifications evaluated with either analytical~\cite{Fast2018, Srinivasan2021} or domain experts~\cite{Rajashekar2024, Gu2024} whereas contradicting findings were based on lay-people~\cite{Ma2025, He2023}. Domain experts, regardless of their analytical skills always wanted CXAI systems to verbally explain statistical processes (e.g. how it predicted a patient's outcome) to decide on the validity of the prediction~\cite{Rajashekar2024, Gu2024}. Analytical experts wanted control over defining every step in the analysis (e.g. selecting predictors, visualisation type, model type, analysis pipelines) and only wanted CXAI systems to execute their commands rather than having to program themselves (e.g. in a Jupyter notebook) or interacting with a system (e.g. Tableau)~\cite{Fast2018, Srinivasan2021, Crovari2021, Li2023}. The difference between wanting explanations and wanting complete control could also stem from the type of tasks.Prognostic tasks were associated with users wanting only explanations~\cite{Rajashekar2024, Gu2024}, while diagnostic tasks corresponded with users wanting control~\cite{Rajashekar2024, Gu2024, Li2023, Crovari2021}.

On the other hand, studies showing that users only wanted explanations were done on crowd workers~\cite{Ma2025, He2023}. In these studies, users did ask CXAI systems to provide evidence for recommended decisions (e.g. why a student should be admitted to university) but did ask how the system reached this recommendation. In non-data analysis tasks (e.g. looking up information about a chosen topic), users again did not ask CXAI systems to elaborate on sources, how the system searched for information, or how it aggregated it~\cite{Wang2024, Poser2022, Brachten2020}. While CXAI systems should still aim for transparency towards all users and not just experts, expert users demand the explainability before trusting the system as they either know the impact that different statistical methods have on results or they know that their decisions have large impacts on other people. Novice users on the other hand do not require such explainability, especially not when presented in a uninterpretable way (e.g. SHAP values) that need expertise to understand. Therefore, CXAI systems for novice users should find simplified methods for explainability to increase transparency and avoid over-reliance.

\section{Discussion}
In this review, we proposed possible factors relating to users and tasks that can distinguish AI user studies from another. We drew on frameworks and studies from multiple fields including XAI, CAI, data analysis, data visualisation, mental workload, and user needs in data analysis tools. We aimed to categorise these studies to explain the inconsistencies between their results, such as those regarding automation bias and global explanations. In this section, we discuss concrete steps towards conceptualising and reporting studies on XAI interfaces.

\subsection{Specifying Users' Expertise}
We hypothesised that many of the inconsistencies stem from users in different studies having different levels of knowledge. For example, using participants from Amazon Turk as opposed to data scientists. One study supports our hypothesis by comparing domain experts with crowd workers~\cite{Wang2024} and other studies have reached similar hypotheses to explain why their own results do not match other studies~\cite{Ahn2024}. Therefore, we propose that studies should specifically report on their users' domain, AI, and data analysis expertise to illustrate the generalisability of their findings. While many studies do report some of their participants' expertise, mainly domain~\cite{Salimzadeh2024} and AI knowledge~\cite{Kaur2020}, most do not report everything needed to accurately compare studies and understand their results. Moreover, most studies only measure users' subjective level of knowledge (i.e. how much the users ranked their own knowledge)~\cite{Salimzadeh2024}, which may not reflect their actual knowledge. Therefore, studies should also report more objective indicators of knowledge. For example, to illustrate domain knowledge, studies can report of education, years of experience, role, or any relevant certifications.

Due to the large impact that users' knowledge has on results, we also recommend that studies replicate previous studies but with different target groups. The vast majority of XAI and CAI studies recruit from non-experts in their studies~\cite{Warren2024, Ma2025, Salimzadeh2024, Wang2021}. This means that systems targetting domain experts (e.g. clinicians, educators, social workers, etc) base their designs based on findings that may not apply to them. Moreover, a large number of studies recruit crowd workers, which may not reflect how users would behave in real use cases. While crowd workers can offer large amounts of data, they often have the goal of completing tasks quickly for financial gain. This means that compared to other users, crowd workers may have overall fewer and quicker interactions with a system compared to non-crowd workers~\cite{Wang2024}.

\subsection{Distinguishing tasks in studies}
Data analysis and visualisation studies have very well defined categorisations of tasks~\cite{munzner2015visualization, Dimara2022, Mndez2017}. However, very few XAI studies use these categorisations to define and distinguish their studies which impact how much we can generalise and compare their results. For example, many non-AI studies use public datasets, publish their datasets, or define their content (e.g. number of variables, data types, number of observations, etc) so that readers can understand the scope of the task~\cite{Mndez2017}. Reporting such details in XAI studies enhances our understanding of task difficulty and complexity, enabling more accurate interpretation of user behaviour and more meaningful comparison across studies. Moreover, XAI studies should also rely on the data visualisation to define their tasks (e.g. exploratory vs confirmatory analysis, diagnostic vs prognostic) for the same reasons. For example, using prior frameworks for understanind data analysis tasks~\cite{munzner2015visualization, Mndez2017, Tukey1980}. The lack of such details in studies makes if more difficult for researchers and designers to understand the current scope of research, what gaps exist, and what design implications apply to their specific cases. 

\subsection{Understanding decision making}
Many studies on data analytics try to define specific steps, needs, and struggle during decision making~\cite{Dimara2022, sacha2015role, WhitelockWainwright2022}. For example, decision making while navigating uncertainty in data due to missing values or data not collected at all~\cite{sacha2015role}. Specifically, decision making can differ greatly between confirmatory and exploratory data analysis~\cite{Tukey1980, Fife2022}. Many XAI studies support exploratory analysis as systems present users with recommendations and explanations before users have a chance to formulate their own opinions~\cite{Sivaraman2023, Wang2021, Bach2023}. However, this limits our understanding of how users negotiate their hypotheses in confirmatory with XAI explanations~\cite{Sivaraman2023}. In CXAI systems for domain experts who are also analysis novices, users should have the ability to define their hypothesis to align their goals with the action recommendations of the system. That way, the system can recommend actions that work towards relevant findings, or even outright present relevant findings.

\subsection{Task realism}
Reviews and studies on mental workload highlight the importance of task realism for accurately understanding user mental states in studies~\cite{Longo2022}. However, many XAI studies investigated tasks with crowd workers that also do not match an identified use case for AI that these users have~\cite{Ahn2024, Linder2021, Solomon2014, Buinca2021, Ma2025}. For example, this can occur in studies with users deciding on college admissions despite having no such role in real life~\cite{Ma2025} or identifying the gender of people in a photo~\cite{Chiang2022}. While these tasks do occur in real life and there is a need for such systems, studies include participants that have no stake (e.g. not working in the field or have never needed such a system). Such studies do not follow the human-centred design principles of investigating a context and identifying user needs~\cite{Harte2017}. When users have not stake, interest, or experience in task, they exert less effort and thus do not reflect how the actual target groups would behave~\cite{Longo2022}. Studies with the highest acceptance of their systems and most consistent findings typically also involve very realistic tasks that matches their users' real tasks as closely as possible~\cite{Bach2023, vanBerkel2022, Zhang2024}. Therefore, future studies should also similarly create more realistic tasks fitting the context and parameters of a real world task they aim to assist with AI.

\subsection{Limitations}
In this study, we propose may dimensions that distinguish studies. However, we do not provide examples for how these dimensions (e.g. categorical v.s. ordered data) can help untangle inconsistent results from different studies. This stems from two reasons. The first being that many studies did not adequately report on what their datasets consisted of~\cite{Salimzadeh2024, Yu2020}, making it difficult to compare them to studies that did~\cite{Warren2024} or those that used publicly available data~\cite{Fast2018, Srinivasan2021, Wang2021}. However, we included these differences in our paper either because some studies did investigate whether they made a difference in user behaviour or because they do impact user behaviour in non-XAI studies (e.g. studies on dashboards). Therefore, future studies still need to investigate how exactly the different task types defined our paper impact user behaviour.

While we present many categorisations to distinguish XAI papers, we also still lack objective measurements or identifying qualities for some. For example, users' domain, AI, and analytical expertise. While we propose proxy measures such as years of experience and certifications, these measures may not accurately represent users' real level of knowledge. Therefore, future studies also need to propose and iterate on our proposed method to further improve our ability to compare studies.

\section{Conclusion}
In this paper, we drew from several fields including AI, data analytics, and psychology to propose a method for categorising and comparing XAI studies under three dimensions: what, why, and who. We identified the main problems being: inadequate descriptions of tasks, context-free studies, and insufficient testing with target users. We specifically propose this categorisation and demonstrate its usefulness in Reconciling existing contradictions in research regarding automation bias, global explanations, and level of explainability in conversational AI. Furthermore, we propose study guidelines for designing their tasks and reporting their work to improve the XAI research community's ability to parse the rapidly growing XAI field. We hope that our framework can help researchers and designers better identify which studies are most relevant to their work and what gaps currently exist in the research.

\section{Acknowledgments}
This work was partially supported by the COST Action IRENE (CA18118) and funded by the Horizon Europe research and innovation programme project RES-Q plus (Grant Agreement No. 101057603). At various points in the paper we used ChatGPT (gpt-4o) to correct grammar mistakes and improve the formatting of tables but no generative AI was used at any point to generate content for the paper.

\bibliographystyle{unsrt}  
%\bibliography{references}  %%% Remove comment to use the external .bib file (using bibtex).
%%% and comment out the ``thebibliography'' section.

%%% Comment out this section when you \bibliography{references} is enabled.

\bibliography{references}

\end{document}